\documentclass[]{article}
\usepackage[letterpaper]{geometry}
\usepackage{amta2020}
\usepackage{times}
\usepackage{url}
\usepackage{latexsym}
\usepackage{natbib}
\usepackage{layout}
\usepackage{graphicx}

\begin{document}

\title{\bf On Target Segmentation for \\ Direct Speech Translation}

\author{\name{\bf Mattia A. Di Gangi}\thanks{The first author performed this work while he was a Ph.D. student at FBK.} \hfill  \addr{mdigangi@apptek.com}\\
\addr{AppTek GmbH, 52062 Aachen, Germany}\\
\AND
\name{\bf Marco Gaido} \hfill \addr{mgaido@fbk.eu}\\
\name{\bf Matteo Negri} \hfill \addr{negri@fbk.eu}\\
\name{\bf Marco Turchi} \hfill \addr{turchi@fbk.eu}\\
\addr{Fondazione Bruno Kessler, Trento, Italy} \\
}

\maketitle
\pagestyle{empty}

\begin{abstract}
  Recent studies on direct speech translation show continuous improvements by means of data augmentation techniques and bigger deep learning models. While these methods are helping to close the gap between this new approach and the more traditional cascaded one, there are many incongruities among different studies that make it difficult to assess the state of the art. Surprisingly, one point of discussion is the segmentation of the target text. Character-level segmentation has been initially proposed to obtain an open vocabulary, but it results on long sequences and long training time. Then, subword-level segmentation became the state of the art in neural machine translation as it produces shorter sequences that reduce the training time, while being superior to word-level models. As such, recent works
  on speech translation started using target subwords despite the initial use of characters and some recent claims of better results at the character level. In this work, we perform an extensive comparison of the two methods on three benchmarks covering 8 language directions and multilingual training. Subword-level segmentation compares favorably in all settings, outperforming its character-level counterpart in a range of 1 to 3 BLEU points. 
\end{abstract}

\section{Introduction}

The recent surge in direct (or end-to-end) speech translation (ST) research \citep{berard2016listen,weiss2017sequence} has favored a blooming of models and data augmentation techniques that contributed to increase the translation quality scores. However, despite the evidence of a general improving field, the state of the art is still not clearly defined and some contrasting claims are present in literature. One aspect that still needs clarification is the segmentation of the words in the target sentences. Some studies followed the setting of Listen, Attend and Spell \citep{chan2016listen} for automatic speech recognition (ASR) and segmented the target text at character level \citep{weiss2017sequence,berard2018end,digangi2019adapting}. 
Other studies followed 
\cite{jia2019} in using subword-level segmentation \citep{pino2019harnessing,bahar2019comparative,liu2019end}, motivated by the impractical training time when using a character-level segmentation on large datasets due to the increased sequence length. 

A research strand in neural machine translation (NMT)  \citep{cherry-etal-2018-revisiting,kreutzerlearning,ataman-etal-2019-importance}, based on recurrent models, showed that character-level models can outperform subword-level models, but need architectural changes to make the computation more efficient and increase the capacity of the models. Indeed, while the subword-level models can store, to some extent, syntactic and semantic information of words in their embedding layers, character embeddings cannot do the same as the function of a word is not obtained by character composition. 

The idea of the superiority of the subword level is suggested for speech translation by 
\cite{bahar2019using}, who claim that their baseline is stronger than the ones used in previous studies because of their use of BPE-segmented words \citep{sennrich2015neural} on the target side instead of characters.
In contrast, two recent works in multilingual speech translation claim better results when using characters, and consequently do not show the weaker results when using BPE \citep{digangi2019one,inaguma_hirofumi_2019_3525560}.
To the best of our knowledge, the only previous study that uses both segmentation strategies was proposed by 
\cite{indurthi2019data}. However, the systems built for the two settings are not directly comparable  with each other as they use different data and training scheme. In particular, they use data augmentation for subword-level models but not for character-level models.
By itself, data augmentation can be sufficient to explain the higher scores in the former case, hiding the contribution of the different segmentation method.
In this paper, we aim to assess the effect of target-side segmentation by directly comparing character- and subword-level models on some of the most popular ST benchmarks. 
In particular we use: Augmented Librispeech \citep{kocabiyikoglu2018augmenting}, the most used benchmark so far; MuST-C \citep{mustc19}, which represents the largest training set for 8 language directions; How2 \citep{sanabria18how2}, which is a quite large dataset and has been used in a recent IWSLT shared task \citep{niehues_j_2019_3525578}.
In order to shed light over this contrasting claims we perform our experiments both in a classical one-to-one translation and in a one-to-many multilingual translation settings. Our results show that the subword-level segmentation, in our setting, is always preferable and it outperforms the character-level segmentation by up to 3 BLEU points \citep{papineni2002bleu}. The gap is confirmed in the multilingual setup.
A final analysis shows that subword-level models are superior independently from the sequence length, and their increased capacity allows them to choose among more options in the generation phase.

\section{Direct Speech Translation}
Spoken language translation (hereby speech translation) has been traditionally performed using a cascade of statistical models involving at least ASR and MT \citep{casacuberta2008recent}, which benefits from the possibility of using state-of-the-art models for all the components, but the translations may be negatively impacted by early wrong decisions (error propagation).
Direct speech translation aims to solve the error propagation problem by removing all the intermediate steps. A single sequence-to-sequence model is trained to output the translation of the given audio input and all its parameters are optimized jointly for the translation task by minimizing the cross-entropy loss.
Besides the goal of outperforming the cascaded approach, direct ST can provide additional benefits such as lower latency and computational cost, and a reduced memory footprint.

\paragraph{Model.} We use the S-Transformer architecture \citep{digangi2019adapting}, an adaptation of Transformer \citep{vaswani2017attention} to the ST task. S-Transformer takes as input a spectrogram to which it applies two successive strided 2D convolutional neural networks (CNNs, \citep{lecun1998gradient}), each followed by batch normalization \citep{ioffe2015batch} and ReLU nonlinearities. The output of the second CNN block, whose size is one fourth of the original size in both dimensions, is then processed by two successive 2D self-attention layers \citep{dong2018speech}. The output is projected to the Transformer encoder layer size and summed with the positional encoding before being passed to the self-attention layers. The Transformer encoder and decoder are equal to the original Transformer, except for a logarithmic distance penalty used in the encoder multi-head attention layers to focus the attention on a local context. 

\section{Target-side segmentation}
\label{sec:hypo}
Character and subword segmentation are two different approaches to achieve an open vocabulary in NLP models. However, their differences directly impact the number of model parameters, the sequence lengths and the re-use of a model.

\paragraph{Characters.} Early work in direct ST splits target text into characters \citep{weiss2017sequence,berard2018end}, following \cite{chan2016listen} that presents the first promising sequence-to-sequence models for ASR. 
Character-level segmentation is appealing because it produces a small dictionary that directly results in a small number of embedding parameters. Also, it removes preprocessing steps and additional hyperparameters that contribute to the \textit{pipeline jungle} that accounts for technical debt in deployed systems \citep{sculley2015hidden}. On the other side, character-level segmentation results in very long sequences that affect negatively the training time, while the dependency range between words becomes longer, hence more difficult to learn.

\paragraph{Subwords.} Subword-level segmentation uses algorithms that split all the words of a language in subwords according to rules that are motivated linguistically or statistically. The most popular segmentation algorithm in MT is the byte-pair encoding (BPE, \citep{sennrich2015neural}), which learns $N$ rules for creating tokens in an iterative fashion. The initial set of tokens is given by the characters in the vocabulary. Then, for $N$ iterations, the most frequent two-token sequence is merged in a new token that is added to the current set, and a new rule is created to generate the new token from the constituents. $N$ is a hyperparameter of the model, and all the data processed by the model have to be segmented using the same learnt rules. Subword-level segmentation allows to find a trade-off between sequence length and dictionary size and reduce the token sparseness. With small data, a linguistically-motivated segmentation  provides a strong inductive bias \citep{ataman2017linguistically} that prevents the emerging of implausible subwords.


\paragraph{Comparison.} In MT, despite some early attempts to use character-level NMT \citep{lee-etal-2017-fully}, it soon became clear that this type of models requires some additional precautions to be competitive with BPE-based models \citep{cherry-etal-2018-revisiting,kreutzerlearning,ataman-etal-2019-importance}. The internal sequence representation should be compressed horizontally in order to reduce both the dependency range and the computational burden. However, this can be done on the encoder side and not in the decoder side, where the compression cannot happen until the new tokens have been generated.
At the same time, \cite{cherry-etal-2018-revisiting} suggest that the embedding layer for characters cannot store useful information, as characters do not have a semantic or syntactical use, and thus more layers and more capacity is required for the whole model. 
Indeed, the model should learn basic character sequences that form words and how they interact. This process requires a large capacity to memorize sequences.
In the rest of this paper, we compare models trained using the two types of target segmentation in a hyperparameter setting that we found good for the character-level models. Our experimental results show a consistent higher quality for subword-level models, and a following analysis will provide insights about the differences in the two systems.

\section{Experimental setting}

\subsection{Model and training settings.}
\label{ssec:setting}
We use the S-Transformer BIG+LOG \citep{digangi2019adapting} setting, which uses convolutions with stride (2, 2), $16$ output channels in each convolution, Transformer layers of size $512$ and hidden feed forward layer size of $1,024$. All our models are  trained with the adam optimizer \citep{kingma2014adam} using the policy proposed in \citep{vaswani2017attention} and dropout 0.1. We always use 4,000 warm-up training steps, and final learning rate of $5e^{-3}$. The loss to optimize is cross entropy with label smoothing \citep{szegedy2016rethinking} set to 0.1. The maximum batch size, where not otherwise reported, is given by the minimum of 768,000 audio frames or 512 segment pairs, achieved using delayed updates \citep{saunders2018multi}.
Sequences longer than $2,000$ audio frames (20 seconds) are discarded from the training set to prevent memory errors.
Decoding is performed with beam search of size 5. All trainings are performed on two GPUs Nvidia K80 with 12G of RAM. The systems are implemented using the codebase in url{https://github.com/mattiadg/FBK-Fairseq-ST}, which is based on Fairseq \citep{gehring2017convolutional} and hence PyTorch \citep{paszke2017automatic}.

For all settings, we first train an ASR model on the (\textit{audio}, \textit{transcript}) portion of the dataset, and then use its encoder weights to initialize the encoder of our ST models.

\subsection{Datasets}

\begin{table}[ht]
\centering
\begin{tabular}{l|llll}
Dataset & Hours & \# Pairs & \# Char & \# BPE \\\hline
Librispeech       & 200 & 94.5K & 132  & 8154        \\
How2              & 300 & 185k  & 124  & 8196         \\\hline
\multicolumn{5}{c}{MuST-C}                     \\\hline
En-De             & 408 & 234k  & 188  & 7436  \\
En-Es             & 504 & 270k  & 348  & 7828  \\
En-Fr             & 492 & 280k  & 212  & 7268  \\
En-It             & 465 & 258k  & 180  & 7500  \\
En-Nl             & 442 & 253k  & 180  & 7444  \\
En-Pt             & 385 & 211k  & 180  & 7532  \\
En-Ro             & 432 & 240k  & 204  & 7676  \\
En-Ru             & 489 & 270k  & 244  & 7468  \\
\end{tabular}
\caption{Statistics of all the corpora. BPE vocabulary is computed with 8,000 merge rules.}
\label{tab:datasets}
\end{table}

We use three speech translation datasets: Augmented Librispeech \citep{kocabiyikoglu2018augmenting}, How2 \citep{sanabria18how2} and MuST-C \citep{mustc19}. They vary in terms of size, language and domain, allowing us to experiment in different conditions. Their statistics are summarized in Table \ref{tab:datasets}.

\noindent
\textbf{Augmented Librispeech} is a small English$\rightarrow$French dataset that has been widely used in previous studies. It is built starting from the English audiobooks of Librispeech \citep{panayotov2015librispeech}, whose texts have been aligned with the French translations of their books. The result is a corpus of 100 hours of speech and 47k segments, which has been doubled by translating all the source segments with Google Translate.

\noindent
\textbf{How2} is an English$\rightarrow$Portuguese corpus for multimedia translation. It has been built from video-tutorials downloaded from Youtube with their transcripts and then translated into Portuguese. Its size is 50\% larger than Augmented Librispeech.

\noindent
\textbf{MuST-C} is a one-to-many multilingual corpus for English$\rightarrow${German, Spanish, French, Italian, Dutch, Portuguese, Romanian, Russian}. It has been built from TED talks, using transcripts and translations provided by \url{www.ted.com} after some steps of filtering. 
Every portion of MuST-C is currently the largest ST corpus available for the corresponding language pair.

\subsection{Data processing and evaluation}
For all audio sequences, we compute their spectrograms using 40 Mel-filterbanks. The audio frames are extracted with windows of $25$ milliseconds and steps of $10$ milliseconds. Only for Librispeech, an additional energy feature is extracted, as described in \citep{berard2018end}.

The target texts are tokenized (and special characters deescaped) using the Moses toolkit \citep{koehn2007moses} tokenizer. Text between parentheses is removed from the training data and from the references in order to remove non-speech audio effects.\footnote{MuST-C contains annotations like (Applause), (Laughter), (Music), and others.} Finally, the resulting texts are split either into characters as in \citep{digangi2019adapting} or with BPE \cite{sennrich2015neural}\footnote{We used the implementation provided in https://github.com/rsennrich/subword-nmt}. The unidirectional translation experiments use $8,000$ merge rules\footnote{It performed better than alternatives in preliminary experiments, but we observed little variability in results.}, while the multilingual experiments compute $20,000$ merge rules among all the target languages. 
English texts, used for training the ASR models, are lower-cased and their punctuation is removed before word segmentation.

All results are computed using case-sensitive BLEU score on word-level text tokenized with the Moses toolkit and evaluated with \texttt{multi-bleu.pl}, unless otherwise specified.

\section{Experiments and Results}
In this section we report on two blocks of experiments. In the first block, we train separate unidirectional ST models on Librispeech, each language portion of MuST-C, and How2, analysing how the two segmentation methods affect the final results.
In the second block of experiments, we run one-to-many-multilingual training on MuST-C using two different approaches, again with both segmentation techniques. 
For all experiments, we use the same set of hyperparameters described in \S \ref{ssec:setting}.

\begin{table}[h]
\centering
\small
\begin{tabular}{l|ll}
Work & Notes & BLEU  \\\hline
\citep{berard2018end} & Multitask & 13.4 \\
\citep{digangi2019adapting} & 13.5 \\
\citep{liu2019end} & PT & 14.3 \\
\citep{inaguma_hirofumi_2019_3525560} & BIG + SP & 15.7 \\
\citep{pino2019harnessing} & PT + TTS PT & 16.4    \\
\citep{liu2019end} & KD & 17.0 \\
\citep{bahar2019using} & BIG + PT + SA & 17.0   \\
\citep{pino2019harnessing} & BIG + MT & 21.7 \\\hline
This work Char                & PT & 16.2 (16.5) \\
This work BPE                & PT  & 17.1 (17.2)
\end{tabular}
\caption{Results on Librispeech. Notes: Multitask = Multitask training with ASR; PT = pretraining; SP = speed perturbation; TTS PT = pretraining on synthetic data generated with TTS; KD = knowledge distillation; SA = SpecAugment; BIG = large model.}
\label{tab:librispeech}
\end{table}

\begin{table*}[t]
\centering
\small
\begin{tabular}{l|llllllll}
Model & De   & Es   & Fr   & It   & Nl   & Pt   & Ro   & Ru   \\\hline
\multicolumn{9}{c}{\citep{indurthi2019data}} \\\hline
Metalearning Char & 17.2 & - & 29.2 & - & - & - & - & - \\
Cascade BPE & 20.9 & - & 33.7 & - & - & - & - & - \\
Metalearning BPE & 22.1 & - & 34.1 & - & - & - & - & - \\\hline
\multicolumn{9}{c}{\citep{pino2019harnessing}} \\\hline
Cascade & - & - & - & - & - & - & 21.0 & - \\
Direct & - & - & - & - & - & - & 17.3 & - \\\hline
\multicolumn{9}{c}{\citep{nguyen2019trac}} \\\hline
MuST-C only & - & - & - & - & - & 23.6 & - & - \\
+ How2      & - & - & - & - & - & 26.9 & - & - \\\hline
\multicolumn{9}{c}{\citep{digangi2020instance}} \\\hline
MuST-C only & 17.0 & 21.5 & 27.0 & 17.5 & 21.8 & 21.5 & 16.4 & 12.2 \\\hline
\multicolumn{9}{c}{This work} \\\hline
Char  & 17.6 & 21.6 & 26.5 & 18.0 & 21.5 & 21.6 & 16.9 & 11.6 \\
BPE   & \textbf{19.1} & \textbf{23.7} & \textbf{30.3} & \textbf{20.1} & \textbf{23.1} & \textbf{23.7} & \textbf{19.5} & \textbf{12.8} \\\hline
\end{tabular}
\caption{Results on MuST-C test sets.}
\label{tab:must-c}
\end{table*}

\subsection{Unidirectional direct ST}

\paragraph{Librispeech.} Table \ref{tab:librispeech} shows the results for Librispeech and compares them with previous studies. Our character-level model outperforms the one in  \citep{digangi2019adapting} by $+2.7$ BLEU points, probably due to the different learning policy with a higher learning rate,\footnote{The work by Di Gangi and colleagues used Adam with a fixed learning rate of $1e^{-4}$ for Librispeech.} and achieves a score of $16.2$. The BPE-level model further improves this score by additional $0.9$ points, reaching the score of $17.1$. Among the other results, we are able to compare  only with the work by Bahar and colleagues, who used \texttt{mteval-v13.pl} for their evaluation. Our results computed with the same script are reported in parentheses. We find remarkable that our BPE model obtains the same score as the one reported in \cite{bahar2019using}, which has been trained with SpecAugment \citep{park2019specaugment}. The other results have been added, despite not being comparable, to show the range of the published scores. Notice that, in \citep{pino2019harnessing}, the authors achieve only 16.4 BLEU when augmenting the dataset with synthetic audio and encoder pre-training. To improve this  score up to 21.7, the same authors used a big Transformer model trained on synthetic parallel data generated from the ASR Librispeech dataset translated with an NMT system. Table \ref{tab:librispeech} also lists the augmentation methods used by the other studies.
%
%
%
%
%
%
These results highlight the importance of the training hyper-parameters. In particular, our model has about 35M parameters, and its results are comparable with the models used in most of the other studies that have almost 300M parameters, and are probably overfitting the training data.

\paragraph{MuST-C.} The results on all the languages of MuST-C are presented in Table \ref{tab:must-c}. Our character-level results are similar but not identical to the ones presented in \citep{digangi2020instance}. Our BPE-level results outperform the ones at character level by at least 1.2 BLEU point on En-Ru and up to 3.3 points on En-Fr, with improvements of about 2 points in most of the languages. 
This difference should be taken into account when comparing with results reported in literature, which are  obtained with a different segmentation technique.

\noindent
Also for this benchmark, we report indicative results from previous work but we cannot ensure that they are obtained with the same tokenization.
However, the scores of our models are in the same range as in the previous studies, confirming their value as strong baselines.

\begin{table}[h]
\centering
\begin{tabular}{l|l}
Model &  BLEU  \\\hline
\citep{digangi2020instance} & 39.4 \\
 + MuST-C                  & 41.0 \\\hline
\citep{nguyen2019trac} CHAR & 39.9 \\
 + MuST-C BPE              & 43.8 \\\hline
This work Char             & 39.1 \\
This work BPE              & 41.2 \\
 + MuST-C                  & 43.5
\end{tabular}
\caption{Comparing our results on How2 with the published results.}
\label{tab:how2}
\end{table}

\paragraph{How2.} For this experiment, we trained two models with the two different segmentation techniques, and additionally we trained another BPE-based model concatenating the two training sets of MuST-C and How2. The results are presented in Table \ref{tab:how2}. The performance of our character-level model is slightly worse but comparable with the results reported in \citep{digangi2020instance} and in 
\citep{nguyen2019trac}. 
Both models are trained with character segmentation, but the second one used a training set augmented with speed perturbation. Our subword-level model outperforms significantly all the character-level models, obtaining a result of $41.2$ that is equivalent to the score 
reported in
\cite{digangi2019adapting} 
when using also MuST-C En-Pt. If we also add MuST-C to the training set, we get a comparable score to 
\cite{nguyen2019trac} in the same condition (but using only 400 BPE rules, the best number of segmentation rules according to their experiments), which additionally used speed perturbation for data augmentation.


\begin{table}[]
\small
\centering

\begin{tabular}{l|ll|llll}
              & De   & Nl   & Es   & Fr   & It   & Pt   \\\hline
\multicolumn{7}{c}{Char} \\\hline
Unidirectional    & 17.6 & 21.5 & 21.6 & 26.5 & 18.0  & 21.6 \\\hline
C-Decoder   & 18.4 & 21.9 & 21.1 & 25.4 & 17.4 & 21.5 \\
S-Decoder   & 18.2 & 21.7 & 21.9 & 26.5 & 18.5 & 22.7 \\\hline
\multicolumn{7}{c}{BPE} \\\hline
Unidirectional     & 19.1 & 23.1 & 23.7 & 30.3 & 20.1 & 23.7 \\\hline
C-Decoder    & 20.7 & 24.6 & 22.6 & 27.6 & 19.3 & 23.3 \\
S-Decoder   & 20.9 & 25.4 & 23.5 & 29.9 & 20.1  & 25.1 \\\hline
\end{tabular}
\caption{BLEU score results with  \textit{concat} (C-*) and \textit{sum} (S-*) \textit{target forcing} on $6$ languages.
Unidirectional refers to \textit{one-to-one} systems.
The other results are computed with one multilingual system for En$\rightarrow${De,NL} and one for En$\rightarrow${Es,Fr,It,Pt}.}

\label{tab:multilingual}
\end{table}

\subsection{Multilingual direct ST}
One-to-many multilingual translation uses a single model to translate one source language (English in this case) into two or more target languages. As there is a single encoder and a single decoder shared among all the language directions, the model requires to be informed about the target language. 
To this aim, the \textit{target forcing} approach by \cite{johnson2017google} pre-pends to the source sentence a ``language token'' indicating the target language. 
In multilingual ST there is no source sentence (the input is indeed a sequence of audio features), so some variations have been proposed to 
adapt this approach.
\cite{inaguma_hirofumi_2019_3525560} proposed to prepend the language token to the target sequence, replacing the \textit{start of sentence} token. \cite{digangi2019one} proposed two  
variants: \textit{concat}, which is analogous to the method proposed by Inaguma and colleagues; and \textit{sum}, which performs an element-wise sum of the language token embedding to all the elements in the target sequence. 
Here, we run experiments at both character- and BPE-level with the two methods and verify if their claims about the BPE segmentation being less effective holds true in the multilingual case. 

We train two groups of systems, one for En$\rightarrow$\{De,Nl\} (Germanic) and one for En$\rightarrow$\{Es,Fr,It,Pt\} (Romance), using both segmentation strategies as well as the two \textit{target forcing} variants in the decoder. The results are presented in Table \ref{tab:multilingual}. Our character-level models improve slightly and comparably over the
unidirectional systems for the Germanic targets, while for the Romance targets the \textit{sum} approach is better by improving by more than 1 BLEU point for Portuguese and smaller improvements in the other cases. However, the difference between the \textit{concat} and \textit{sum} approaches is more evident in the BPE-level case, where the former is generally worse than the
unidirectional systems and the latter is comparable or better. The larger improvements are for German, Dutch and Portuguese with, respectively, $+1.8$, $+2.3$ and $+1.4$. For the Romance languages, the \textit{concat} approach produces in general a score degradation with respect to the unidirectional systems. 
The \textit{sum} approach, instead, is at least on par with the unidirectional baselines.
The only real improvement among Romance languages can be measured for Portuguese, which has about 25\% less data than the other languages and can benefit from positive transfer learning.
The reason of the different behaviors of multilingual approaches is beyond the scope of this paper and we defer the analysis to other studies.
These results indicate that, besides significantly outperforming the character-level models also in the multilingual setting (up to $3.7$ BLEU points in En-Nl), under some conditions (for the Germanic languages) the BPE-level models can benefit more also from the multilingual training, widening the already large gap between the baselines.

\section{Analysis}
\begin{figure}[h]
    \centering
    \includegraphics[width=0.5\textwidth]{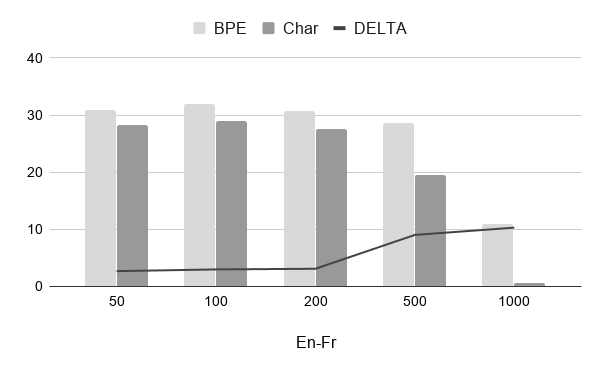}
    \caption{BLEU score by reference length (number of characters) for character- and BPE-level models in English-French.}
    \label{fig:enfr}
\end{figure}

\begin{figure}[h]
    \centering
    \includegraphics[width=0.5\textwidth]{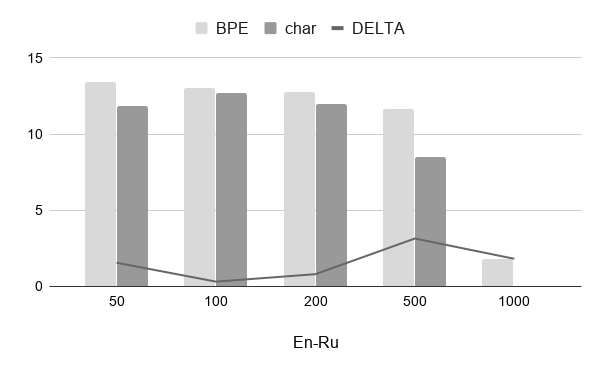}
    \caption{BLEU score by reference length (number of characters) for character- and BPE-level models in English-Russian.}
    \label{fig:enru}
\end{figure}

We are now interested in understanding why the BPE-level models are better performing than their character-level counterparts. To this aim, we analyze
the translation quality on sentences with different target lengths
and the output distribution peakiness. The goals are: \textit{i)} 
 checking if  the differences between the two models depend on the sentence length, 
 and \textit{ii)} verifying 
 the idea that character-level models need to memorize more (see \S \ref{sec:hypo}).
\paragraph{Length comparison.} The plots in Figures \ref{fig:enfr} and \ref{fig:enru} show the BLEU scores achieved by the systems for groups of different reference length (measured in number of characters) for, respectively, En-Fr and En-Ru. Let DELTA be the reference length difference between BPE and character segmentations, we can observe that it has a clear upward trend in En-Fr, while in En-Ru DELTA is larger in the first bin, before increasing again for the longest sentences. The trend for character-level models is to have a larger BLEU score in the second bin than in the first one, then a smooth transition to the third bin, and finally a dramatic degradation in the last two bins. The degradation in the last two bins is less 
evident
for the BPE-level models. Similar patterns are found in the other language directions. This first analysis shows that BPE-level models achieve a higher quality in all length bins and can 
better manage 
long translations.

\paragraph{TER.} The TER \citep{snover2006study} scores translations at sentence level as a function of errors (insertions, deletions, substitutions, shifts) with respect to a reference. We compute TER with a focus on MuST-C French translations to obtain more details about the score difference between the two systems. We use the sentence-level scores to divide the translations in three groups according to the system that produced the best translation (or a tie). We say that the system that produces the best translation is the winner for that segment. The results are summarized in Table \ref{tab:ter}. First, we can observe a big difference on the percentage of wins for the two systems: the BPE-level system wins in 44.6\% of the cases, against 30.2\% for the Char-level system and only a 25\% of ties. When breaking the results according to the segment groups, both systems achieve the lowest average error rate in the Tie group. By a manual inspection, we found that these are not easy or short sentences that are translated equally, but
the systems generate different translations often with different errors, which are overall balanced though.
Then, it is interesting to understand whether a pattern exists in the Char Winner group. We found that these are mostly hard cases in terms of terminology. Both systems produce poor translations in such cases, but the shorter translations by the Char-level system reduce the TER. For the BPE winner segments, the TER difference can range from small to high values, and it increases with the reference length. To summarize, this inspection found that the BPE-level system is equal to or better than the Char-level system on 70\% of the segments, and the superiority is clear on long segments. Also, it suggests that systems trained on such small datasets suffer from significant vocabulary issues and
data augmentation methods can be considered as a partial but effective solution to this problem.

\begin{figure}
    \centering
    \includegraphics[width=0.45\textwidth]{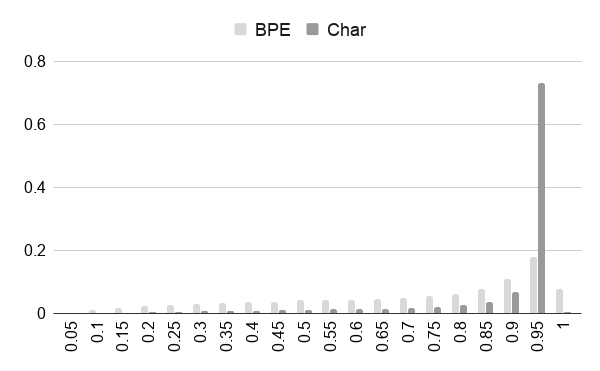}
    \caption{Histograms of the probabilities of the symbols selected by beam search in the test set of MuST-C En-Fr.}
    \label{fig:histogram}
\end{figure}

\begin{table}[h]
\centering
\begin{tabular}{l|l|l|l||ll}
     & Char Winner & BPE Winner & Tie    & Total  & \% Win\\\hline
Char & 51.2       & 61.4      & 46.3  &  57.6   &   30.2  \\
BPE  & 66.1       & 46.4      & 46.3  &  53.9   &   44.6
\end{tabular}
\caption{Average TER of the two systems on the groups of sentences where the best translation is produced by Char, BPE, or there is a tie. The best translation is decided according to the TER score for each individual sentence. The last two columns show the TER computed on the whole test set and the percentage of times when one system won.}
\label{tab:ter}
\end{table}
\paragraph{Distribution peakiness.} The peakiness of the output conditional distribution is an indicator of how the model is confident in its choice.
Over-confident distributions are a signal of over-fitting and \cite{meister-etal-2020-generalized} showed that models producing higher-entropy distributions provide better translation.
Here, we consider the conditional probability of each token selected by beam search in the whole test set, and group them in bins of size 0.05. In the general case, this approach would give little information about the whole distribution, but in this case we find that it is sufficient to discriminate the working of the two types of system. We show only the results with MuST-C En-Fr, but the other models follow a similar behavior. As it is shown in Figure \ref{fig:histogram}, both models have their mode in the bin containing probabilities between 0.9 and 0.95. However, while for the BPE-level model the mode accounts for less than 20\% of the cases, for the character-level model it is more than 70\%. Additionally, the cumulative probability of the bins from $0.8$ to $1.0$ is $0.90$. This shows that the probability mass is concentrated in one single symbol in the vast majority of cases. A manual inspection revealed that the few occurrences of low probabilities for the first choice occur mostly in two cases: when choosing the first character of a word, mostly for content words, or for choosing the first character of a suffix when it decides a word inflection. The network is thus capable of outputting easily some memorized sequences, and its uncertainty increases only in a few critical points. This model overconfidence may reduce the effectiveness of beam search in evaluating alternative options because of the large score difference.

\begin{figure}
    \centering
    \includegraphics[width=0.45\textwidth]{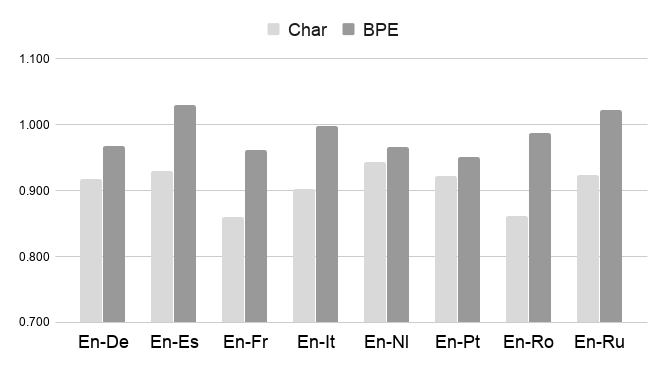}
    \caption{Comparison of the length of the produced output with respect to the reference.}
    \label{fig:len}
\end{figure}

\paragraph{Discussion.} Our analysis revealed that BPE-level models produce, in average, better translations for all reference lengths with a higher difference in the longest sentences.
The inability of character-level models to properly translate long sequences is confirmed by their tendency to generate shorter outputs, as shown in Figure \ref{fig:len}.
Additionally, BPE-level models produce better translations more often than the char-level models do (44\% vs 30\%, see Table \ref{tab:ter}). 
Finally, we found that character-level models generate output conditional distributions that are significantly more peaked, confirming their higher reliance on memorization and overfitting. \cite{cherry-etal-2018-revisiting} observed for NMT that character-level models require a very large capacity, which can be the topic of future work. Here, we showed that our results hold for a large number of domains and target languages, and with strong baseline models. 

\section{Related work}
Direct ST has been proposed with the idea to skip the transcription phase of a cascaded system to improve translation quality \citep{berard2016listen}. However, the small data condition of this task appeared to be the main obstacle to 
overcome, which has
 been tackled 
with techniques like transfer learning or multitask learning \citep{weiss2017sequence,berard2018end} in order to leverage knowledge from the tasks of ASR and MT. This techniques resulted to be useful, but the improvements were limited and really effective only in 
very small data conditions
\citep{anastasopoulos2018tied,bansal2018low}. 
Furthermore, 
\cite{sperber2019attention} argued that current sequence-to-sequence architectures are not effective in leveraging the additional data, and an evolution of the two-stage-decoding model \citep{kano2017structured} is more 
data-efficient. 
In an attempt to transfer knowledge from MT to ST, 
\cite{liu2019end} used knowledge distillation, but the real game-changing approach appeared to be the use of synthetic parallel data generated with TTS and MT systems \citep{jia2019}. \cite{pino2019harnessing} showed that using data with the target side generated by MT outperforms model pretraining. Separately, Bahar and colleagues studied the contributions of pretraining and of the CTC loss for multitask learning \citep{bahar2019comparative} and the effectiveness of SpecAugment \citep{bahar2019using}, but without synthetic parallel data. 
In a different attempt to leverage more data,  \cite{digangi2019one} and \cite{inaguma_hirofumi_2019_3525560} proposed multilingual ST. Despite showing a general direction of improvement, the results from these studies are not directly comparable for several reasons: 
\textit{i)}
the use of characters or subwords in the target side; 
\textit{ii)} the use of different model architectures, with sizes ranging from 9M to about 300M parameters; 
\textit{iii)}
the studies experiment only on few datasets, and the results obtained in small datasets are not always portable to larger datasets.
We tried to
shed light on
some of the mentioned problems by 1) proving that the target segmentation is highly relevant for the final result, 2) using a single architecture and model size that can be compared with literature, and 3) setting strong baselines for 10 common benchmarks.

\section{Conclusion}
The optimal target text segmentation was a source of doubts according to literature on direct speech translation, with studies making strong claims in one sense or another without a clear proof. In this work, we performed a thorough investigation across different ST benchmarks using hyperparameters proposed for the character level, and we found that the BPE segmentation is always preferable both in terms of computational burden and translation quality. Our experimental results define new strong baselines for the chosen benchmarks, often performing similarly to models using data augmentation. 
In light of our findings,
we invite the community to take into account the target segmentation when comparing different systems and to take more care of their baselines to really measure 
improvements 
in 
the field.


\bibliographystyle{apalike}
\bibliography{amta2020}

\end{document}